\pdfoutput=1

\documentclass[11pt]{article}

\usepackage[]{EACL2023}

\usepackage{times}
\usepackage{latexsym}
\usepackage{xcolor}
\usepackage{booktabs}
\usepackage{graphicx}
\usepackage{amssymb}
\usepackage{tcolorbox}
\usepackage{enumitem}
\tcbuselibrary{listings}
\usepackage{tcolorbox}
\usepackage{afterpage}
\usepackage{cellspace}

\usepackage{pbox}

\usepackage{ulem}
\usepackage{xargs}

\newtcblisting{promptbox}[1]{%
  title=#1,
  listing only,
  colback=white!95!gray,
  colframe=gray,
  width=\columnwidth,
  arc=0mm,
  boxrule=0.5mm,
  colbacktitle=white!95!gray,
  coltitle=black,
  fonttitle=\bfseries,
  left=6pt,
  right=6pt,
  top=6pt,
  bottom=6pt,
  listing options={basicstyle=\small\ttfamily, breaklines=true}
}

\usepackage[T1]{fontenc}

\usepackage[utf8]{inputenc}

\usepackage{microtype}

\usepackage{inconsolata}

\usepackage{graphicx}

\definecolor{gpt-4}{HTML}{36454F}
\definecolor{gpt-4-proportional}{HTML}{BD7EBE}
\definecolor{gpt-4-balanced}{HTML}{B2E061}
\definecolor{chatGPT}{HTML}{708090}
\definecolor{chatGPT-balanced}{HTML}{7EB0D5}
\definecolor{chatGPT-proportional}{HTML}{FD7F6F}
\definecolor{mTurk}{HTML}{FFB55A}
\definecolor{green}{HTML}{009901}

\newcommand\cincludegraphics[2][]{\raisebox{-0.3\height}{\includegraphics[#1]{#2}}}

\title{The Parrot Dilemma:\\Human-Labeled vs. LLM-augmented Data in Classification Tasks}

\author{Anders Giovanni Møller \\
  IT University of Copenhagen \\
  \texttt{agmo@itu.dk} \\\And
  Arianna Pera \\
  IT University of Copenhagen \\
  \texttt{arpe@itu.dk} \\\AND
  Jacob Aarup Dalsgaard \\
  IT University of Copenhagen \\
  \texttt{jacd@itu.dk} \\\And
  Luca Maria Aiello \\
  IT University of Copenhagen \\
  \texttt{luai@itu.dk} 
  }

\begin{document}
\maketitle
\begin{abstract}

In the realm of Computational Social Science (CSS), practitioners often navigate complex, low-resource domains and face the costly and time-intensive challenges of acquiring and annotating data. We aim to establish a set of guidelines to address such challenges, comparing the use of human-labeled data with synthetically generated data from GPT-4 and Llama-2 in ten distinct CSS classification tasks of varying complexity. Additionally, we examine the impact of training data sizes on performance. Our findings reveal that models trained on human-labeled data consistently exhibit superior or comparable performance compared to their synthetically augmented counterparts. Nevertheless, synthetic augmentation proves beneficial, particularly in improving performance on rare classes within multi-class tasks. Furthermore, we leverage GPT-4 and Llama-2 for zero-shot classification and find that, while they generally display strong performance, they often fall short when compared to specialized classifiers trained on moderately sized training sets.

\end{abstract}

\section{Introduction}
\label{sec:introduction}

Large Language Models (LLMs), such as \texttt{OpenAI}'s GPT-4~\citep{openai2023gpt4}, have demonstrated impressive \textit{zero-shot} performance across a range of tasks, including code generation, composition of human-like text, and various types of text classification~\citep{bubeck_sparks_2023, zhang2022would, savelka2023unlocking, gilardi2023chatgpt}. However, LLMs are not perfect generalists as they often underperform traditional fine-tuning methods, especially in tasks involving commonsense and logical reasoning~\citep{qin2023chatgpt} or concepts that go beyond their pre-training~\citep{ziems2023can}. Additionally, the deployment of LLMs for downstream tasks is hindered either by their massive size or by the cost and legal limitations of proprietary APIs. Recently, competitive open-source alternatives such as Llama~\citep{touvron2023Llama, touvron2023Llama2}, Mistral~\citep{jiang_mistral_2023}, and Falcon~\citep{penedo2023refinedweb} have emerged, allowing their use at a substantially lower cost compared to proprietary models. However, the training dataset sizes of these open-source models do not match those of their closed-source counterparts, and their performance across tasks remains somewhat uncertain.

As an alternative to zero-shot approaches, researchers have explored the use of LLMs for \textit{annotating} data that can be later used for training smaller, specialized models, thus reducing the notoriously high cost of manual annotation~\citep{wang2021want}. Previous work has primarily focused on using LLMs for zero- or few-shot annotation tasks, reporting that synthetic labels are often of higher quality and cheaper than human annotations~\citep{gilardi2023chatgpt, he2023annollm}. However, zero-shot annotations struggle with complex Computational Social Science (CSS) concepts, exhibiting lower quality and reliability compared to human labelers~\citep{wang2021want, ding2022gpt, zhu2023can}. 

Other work has proposed to mitigate these weaknesses by using LLMs to \textit{augment} human-generated training examples~\citep{sahu2022data} either through text completion of partial examples~\citep{feng2020genaug, bayer2023data} or through generation~\citep{yoo2021gpt3mix, meyer2022we, balkus2022improving, dai2023chataug, guo2023dr}. Research on data augmentation with LLMs is still in its early stages, exhibiting two main limitations. First, different classification experiments with synthetic augmentation produced mixed results; some demonstrated improvements in model performance ~\citep{balkus2022improving} while others observed minimal gains or even negative impacts~\citep{meyer2022we}.
A recent review on the topic contributes to the assessment of an unclear landscape~\citep{ollion_chatgpt_2023}, highlighting that substantially smaller models fine-tuned on human-annotated data often outperform the LLMs. Second, most previous work focuses on benchmarks that tend to be homogeneous in terms of their nature and complexity (e.g., sentiment classification), while disregarding more difficult or low-resource tasks. Overall, the benefits of LLMs-based augmentation are not conclusive, especially when using them for training models for complex and low-resource classification tasks typical in Computational Social Science (CSS) research. Such prevailing uncertainty  generates a dilemma of whether it is best to concentrate more resources into manual data labeling or into artificial augmentation.

This work makes two contributions with the aim of bringing more clarity to this complex landscape.
 
First, with the goal of providing CSS practitioners with a set of actionable guidelines for using LLMs in classification, we experiment with synthetic data augmentation on ten tasks of varying complexity typical of the domain of CSS. Second, we perform a comparative analysis of strategies that incorporate LLMs into classification tasks either as data augmentation tools or as direct predictors. Specifically, we assess how augmenting data with LLMs-generated examples performs compared to manual data annotation. We train our classifiers using incrementally larger datasets derived either from crowdsourced annotations or generated by GPT-4 or Llama-2 70B, one of the best-performing open-source alternatives against closed-source model. We then contrast their performance to the zero-shot abilities of both the LLMs considered. 

\noindent Overall, our work contributes to the current literature with three findings:
\begin{itemize}[noitemsep, nolistsep, leftmargin=*]
\item Synthetic augmentation typically provides little to no improvement in performance compared to models trained on human-generated data for binary tasks or balanced multi-class tasks. Such a finding holds even with small amounts of training data and affirms the high value of human labels.
\item More complex tasks benefit more from LLMs-generated data. In the most challenging tasks considered, both in terms of the number of classes and unbalanced data, we demonstrate that synthetic augmentation enhances model performance, substantially beating crowdsourced data.
\item Zero-shot classification is generally outperformed by specialized models trained on human or synthetic data, challenging the belief that LLMs' strong zero-shot performance is the key to mastering complex classification tasks.
\end{itemize}

\section{Methods} \label{sec:methods}

\begin{table}[t!]
\resizebox{\columnwidth}{!}{%
\begin{tabular}
{@{}ccScccc@{}}
\toprule
Task               & \multicolumn{1}{c}{Non-English} & \pbox{2cm}{\centering Small size} & \pbox{2cm}{\centering Class imbalance} & \pbox{2cm}{\centering Sensitive} & \pbox{2cm}{Num. \\ classes} \\ 
             &  & \cincludegraphics[height=.2in]{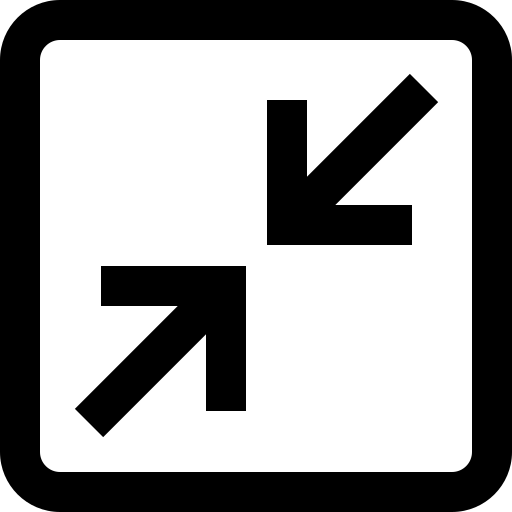} & \cincludegraphics[height=.2in]{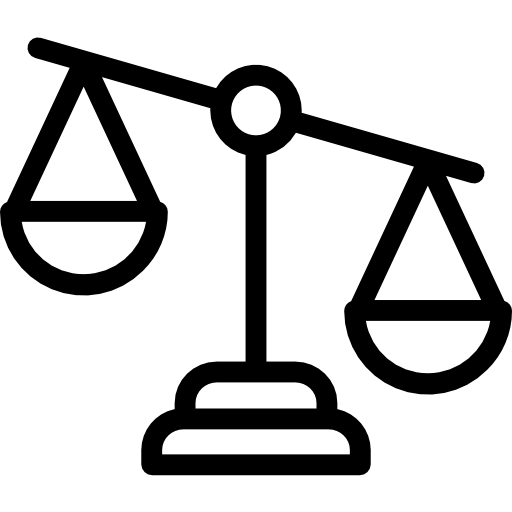} & \cincludegraphics[height=.2in]{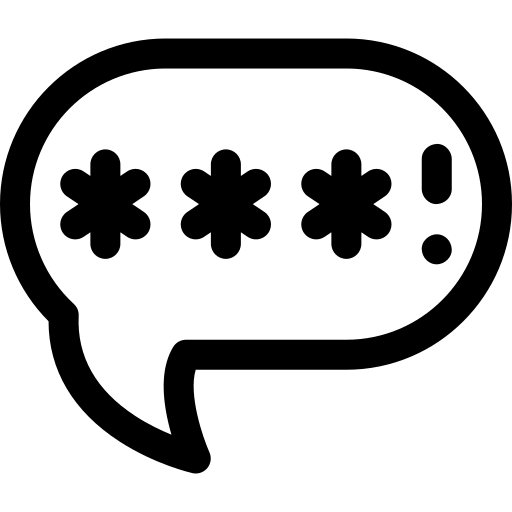} & \cincludegraphics[height=.2in]{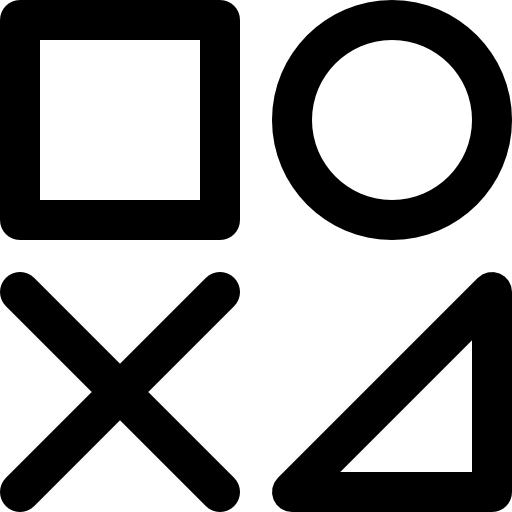} \\ 
\midrule
\texttt{Sentiment} &         &                                             &                 &              &             \multicolumn{1}{c}{2}    \\
\texttt{Offensive} &           \multicolumn{1}{c}{\checkmark}                   &           \multicolumn{1}{c}{}                                &               \multicolumn{1}{c}{\checkmark}         &       \multicolumn{1}{c}{\checkmark}          &        \multicolumn{1}{c}{2}      \\
\texttt{Social dimensions}   &                              &   \multicolumn{1}{c}{}                                          &         \multicolumn{1}{c}{\checkmark}                     &    &    \multicolumn{1}{c}{9}      
  \\
\texttt{Emotions}   &                              &   \multicolumn{1}{c}{}                                          &         \multicolumn{1}{c}{\checkmark}     &                  &    \multicolumn{1}{c}{13}         
  \\
\texttt{Empathy}   &                              &   \multicolumn{1}{c}{}                                          &                             &    &     \multicolumn{1}{c}{2}         
  \\
\texttt{Politeness}   &                              &   \multicolumn{1}{c}{\checkmark}                                          &              & &     \multicolumn{1}{c}{2}         
  \\
\texttt{Hyperbole}   &                              &   \multicolumn{1}{c}{}                                          &                  &   & \multicolumn{1}{c}{2}         
  \\
\texttt{Intimacy}   &                              &   \multicolumn{1}{c}{}                                          &                 & &   \multicolumn{1}{c}{6}         
  \\
\texttt{Same side stance}   &                              &                       \multicolumn{1}{c}{\checkmark}                      &                                     &    &\multicolumn{1}{c}{2}         
  \\
\texttt{Condescension}   &                              &   \multicolumn{1}{c}{}                                          &     &   \multicolumn{1}{c}{\checkmark}    &    \multicolumn{1}{c}{2}  \\       
\bottomrule
\end{tabular}%
}
\caption{\texttt{\textbf{Task properties.}} Characteristics of our tasks in terms of complexity.}
\label{tab:task-difficulties}
\end{table}

We address ten classification tasks within the domain of CSS: (i) \texttt{\textbf{sentiment}} analysis~\citep{rosenthal-etal-2017-semeval}, (ii) \texttt{\textbf{offensive}} language detection in Danish~\citep{sigurbergsson2023offensive}, (iii) extraction of  \texttt{\textbf{social dimensions}} of language~\citep{Choi_2020}, (iv) \texttt{\textbf{emotions}} classification~\citep{CrowdFlower}, (v) presence of \texttt{\textbf{empathy}} in text~\citep{buechel2018modeling}, (vi) identification of \texttt{\textbf{politeness}}~\citep{hayati2021does}, (vii) \texttt{\textbf{hyperbole}} retrieval~\citep{zhang2022mover}, (viii) level of \texttt{\textbf{intimacy}} in online questions~\citep{pei2020quantifying}, (ix) whether two stances are at the \texttt{\textbf{same side}} of an argument~\citep{korner2021classifying}, and (x) detection of \texttt{\textbf{condescension}} on social media posts~\citep{wang2019talkdown}. Data for all tasks is publicly available. Table~\ref{tab:task-difficulties} provides a summary of task difficulties across multiple dimensions.

\begin{figure}[t]
  \centering
  \includegraphics[width=0.48\textwidth]{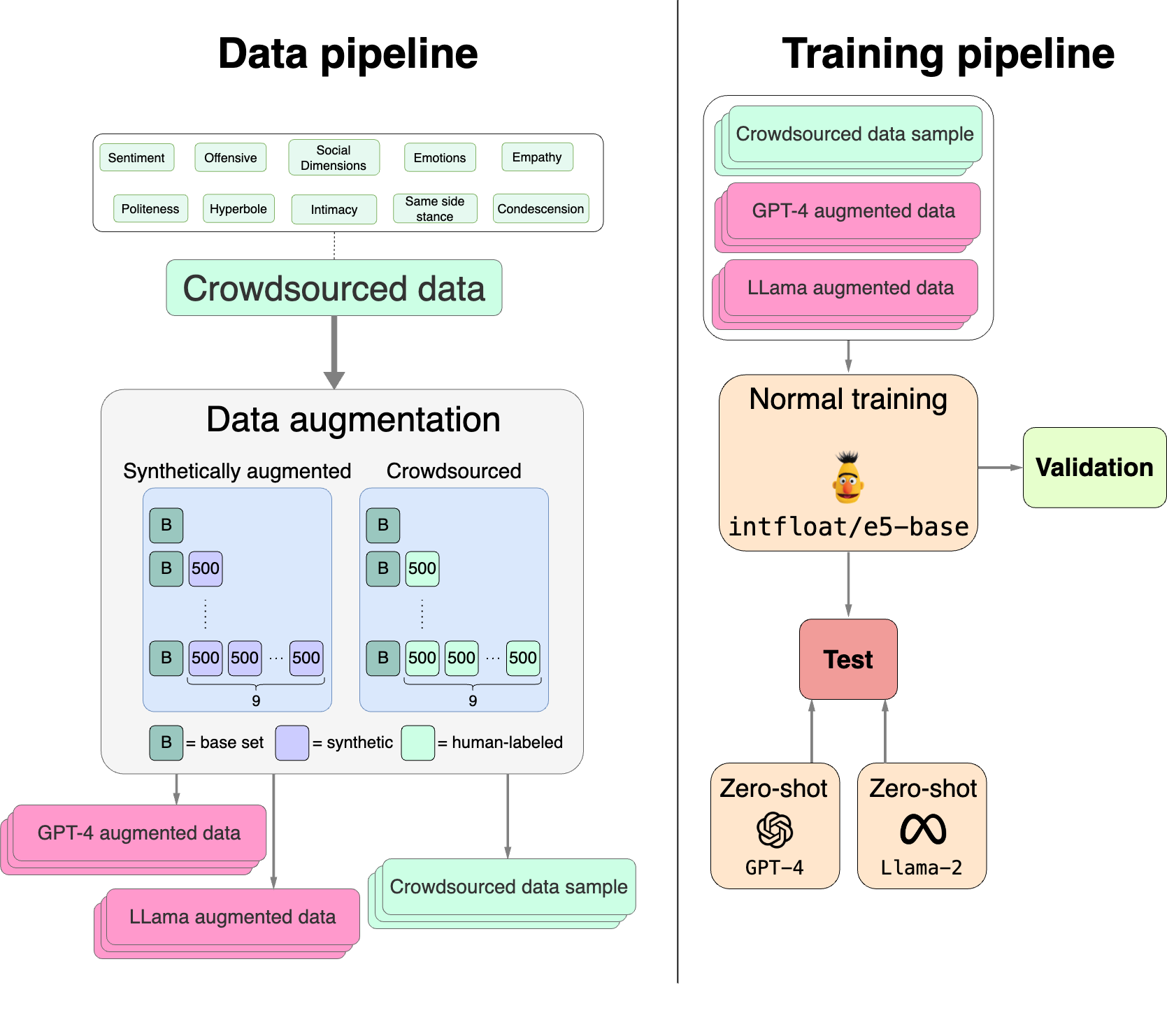}
  \caption{\texttt{\textbf{Experimental framework.}} For each dataset, we start from a base set ($10\%$ crowdsourced samples) and augment it either by adding manually labeled samples or synthetic samples obtained with LLMs. Augmented training sets of different sizes are used to train classifiers. Models are tested on a holdout set and compared to zero-shot approaches.}
  \label{fig:method}
  \vspace{-3mm}
\end{figure}

Our experimental setup simulates a scenario where minimal manually labeled data is available, and additional labels are acquired either through human annotations or synthetic augmentation (Figure~\ref{fig:method}). If test data is already available as separate from the training one in the original sources, we consider such a set as the test set. Otherwise, we reserve 20\% of the original data for testing. Given the diverse sizes of the datasets and the time and economic constraints associated with using LLMs APIs, we have set a threshold of $5,000$ samples to define the \textit{actual training} set. We set aside a fixed base set of $10\%$ samples from the actual training data, which we augment by generating $9$ times the same amount of synthetic texts with GPT-4 and Llama-2 70B Chat~(\S\ref{sec:augmentation}). Subsequently, we construct training sets of increasing sizes, starting from the base set and incrementing by $10\%$ sample size either from the original data (crowdsourced dataset) or the synthetic data (augmented dataset), until reaching a maximum of $100\%$ of the actual training data. For each dataset, we train a separate classifier (\S\ref{sec:classifier}), validate it on $10\%$ randomly sampled data points from the actual training set for each training instance, and evaluate its performance on the holdout test set. To establish a baseline, we compare the trained models' performance with zero-shot classification using GPT-4 and Llama-2 70B Chat. We provide the models with a text and a set of possible labels, requesting them to classify the text accordingly (see Appendix). We use identical prompts for both LLMs, with minimal changes to the template of Llama-2 to align it with its pre-training format. All code and synthetically generated data are available on GitHub\footnote{\url{https://github.com/AndersGiovanni/worker_vs_gpt.git}}.

\subsection{Data Augmentation}
\label{sec:augmentation}

We construct prompts consisting of an example from the original data along with its corresponding label. We instruct the LLMs to generate 9 similar examples with the same label. We adopt a \textit{balanced} augmentation strategy: we first balance the class distribution in the base set by oversampling the minority classes. Then, we augment this modified set by generating 9 examples for each data point. To ensure that the synthetic examples generated from the oversampled classes exhibit substantial differences, we set the temperature to 1. We evaluate the diversity of generated data by examining the cosine similarity (\textit{semantic diversity}, computed with pytorch \texttt{SentenceTransformer}) to the data sample used for the synthetic generation, as well as the fraction of overlapping tokens between the two texts (\textit{lexical diversity}). We provide a detailed explanation of the process in the Appendix.  

\subsection{Classifier training}
\label{sec:classifier}

We use the Huggingface \texttt{Trainer} interface to train \texttt{intfloat/e5-base}~\citep{wang2022text}, a 110M parameter model~\citep{wang_text_2022} that achieves state-of-the-art performance on tasks similar to those we investigate~\citep{muennighoff_mteb_2023}. We train the model in several iterations on the different tasks and datasets. For each iteration, we run the training for $10$ epochs with a batch size of $32$. We use the AdamW~\citep{loshchilov2019decoupled} optimizer with a learning rate of $2e-5$. We track evaluation performance for every epoch iteration. We select the checkpoint with the lowest validation loss and use it to evaluate the test set via macro F1 and accuracy. The runtime for each training instance ranges from $1$ to $31$ minutes. The test performance is overall comparable to the one on the validation set (detail in Supplementary).

\begin{figure*}[t!]
  \centering
  \includegraphics[width=1\textwidth]{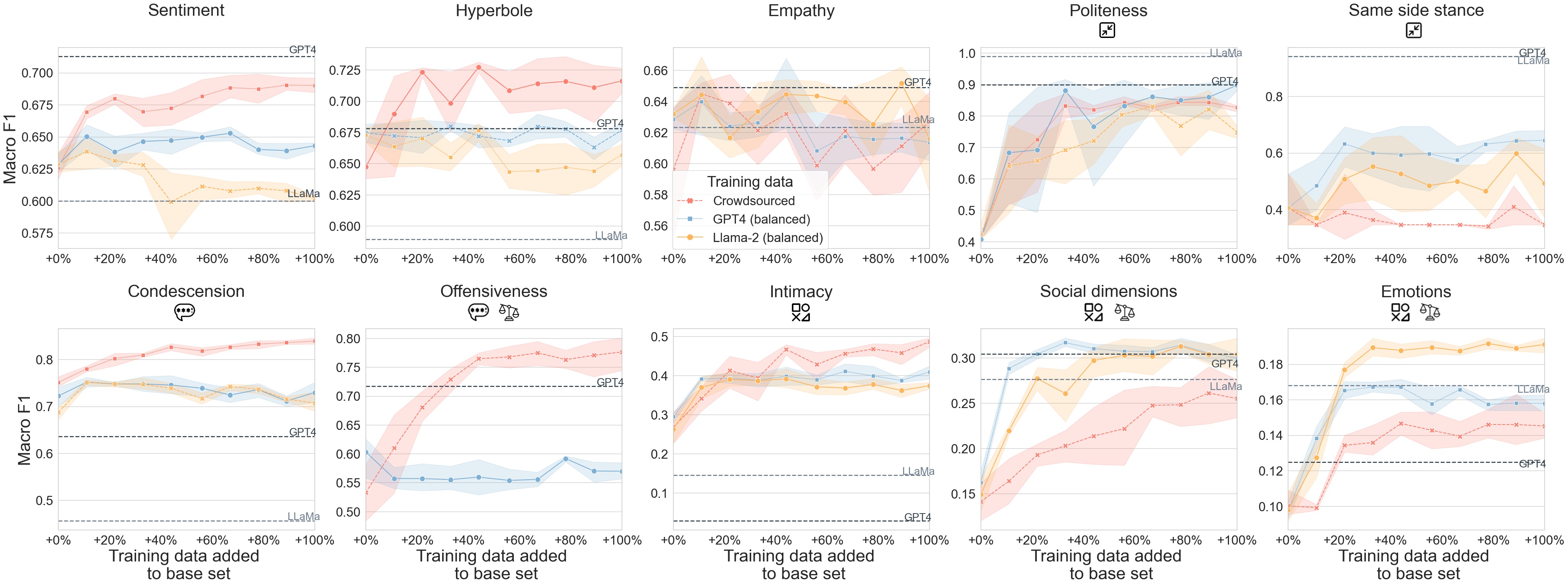}
  \caption{\textbf{\texttt{Data augmentation experiment.}} Macro F1 score on the test set for the ten classification tasks, given various training data sizes and augmentation strategies. Y-axis scales are defined differently for each task to enhance clarity. Each set of training samples contains $10\%$ crowdsourced samples (base set). The dashed line represents the zero-shot performance of LLMs. Each experiment undergoes 5 runs of training with different data sampling seeds and confidence intervals around average metric values are shown. Tasks are grouped by complexity levels (cf. icon tags defined in Table~\ref{tab:task-difficulties}) and sorted within each group by the relative improvement in performance between crowdsourced-based and other types of training.}
  \label{fig:data_size}
  \vspace{-3mm}
\end{figure*}

\section{Results}
\label{sec:results}

Figure~\ref{fig:data_size} illustrates the comparison between classification models trained on varying amounts of human-labeled and synthetically augmented data in terms of Macro F1 score (results for other metrics can be found in Supplementary and on W\&B\footnote{\url{https://wandb.ai/cocoons/crowdsourced_vs_gpt_datasize_v2}}). Three key findings emerge. First, models trained on human-annotated data generally outperform those trained on synthetically augmented data and zero-shot models in the cases of binary balanced tasks (cf. \texttt{hyperbole}), sensitive tasks (cf. \texttt{condescension} and \texttt{offensiveness}) and multi-class balanced tasks (cf. \texttt{intimacy}), even with limited sizes of training data. However, models trained on synthetically augmented data perform well on unbalanced multi-class tasks (cf. \texttt{social dimensions} and \texttt{emotions}), most likely due to the balanced data augmentation technique which substantially increases the number of samples for rare classes. In the specific case of \texttt{emotions}, the classification model based on Llama-2 synthetically generated data outperforms all the other methods. Synthetic data created via Llama-2 is, on average, more diverse from original data than that generated via GPT-4, especially from a lexical perspective (see diversity analysis in the Appendix), which might be beneficial for multi-class unbalanced tasks and particularly for \texttt{emotions}.

Second, zero-shot performance is strong only on specific tasks. 
For GPT-4, this holds particularly for \texttt{sentiment}, likely due to the vast amount of related data in GPT-4's training dataset, and \texttt{same side stance} tasks, possibly because of the small size of the test data available. GPT-4 also performs well in the second smallest dataset considered: \texttt{politeness}. In comparison, Llama-2 performs substantially worse on \texttt{sentiment}, on-par on \texttt{same side stance}, and even better on \texttt{politeness}. For other tasks, the performance of zero-shot models is comparable to or even worse than that of classification models trained on either human-annotated or synthetically augmented data, particularly for \texttt{intimacy} and \texttt{condescension}. Such tasks are characterized by a very nuanced difference between classes and by a notion of social ``power'' that cannot be extracted easily, given the complex paradigm of social pragmatics. A similar case of negative imposition of ``power'' is that of \texttt{offensive}, which is also characterized by a low zero-shot performance likely due to the restrictions of LLMs on offensive language. Overall, only focusing on the zero-shot setting, we observe GPT-4 to be best on six tasks, equal in one task, and Llama-2 best on three tasks. Llama-2 was unable to produce any synthetically augmented text in Danish for the task of \texttt{offensiveness}, thus we decided not to run the zero-shot Llama classification for such a task.

\section{Discussion and Conclusion}
\label{sec:discussion}

To enhance our limited understanding of the ability of LLMs to serve as substitutes or complements to human-generated labels in data annotation tasks, we investigate the effectiveness of generative data augmentation with LLMs on ten classification tasks with varying levels of complexity in the domain of Computational Social Science. Augmentation has minimal impact on classification performance for binary balanced tasks, but shows promising results in complex ones with multiple and rare classes. Our findings lead to three key conclusions. First, the time to replace human annotators with LLMs has yet to come---manual annotation, despite its costliness~\citep{williamson_ethics_2016}, provides more valuable information during training for common binary and balanced tasks compared to the generation of synthetic data augmentations. 

Second, artificial data augmentation can be valuable when encountering extremely rare classes in multi-class scenarios, as finding new examples in real-world data can be challenging. In such cases, our study shows that class-balancing LLMs-based augmentation can enhance the classification performance on rare classes. Lastly, while zero-shot approaches are appealing due to their ability to achieve impressive performance without training, they are often beaten by or comparable to models trained on modest-sized training sets. 
Overall, our study provides additional empirical evidence to inform the ongoing debate about the usefulness of LLMs as annotators and suggests guidelines for CSS practitioners facing classification tasks. To address the persistent inconsistency in results on LLMs' performance, we emphasize two essential requirements: (i) the establishment of a systematic approach for evaluating data quality in the context of LLMs-based data augmentation, particularly when using synthetic samples and (ii), the collaborative development of a standardized way of developing prompts to guide the generation of data using LLMs.

\section*{Limitations}
Constructing a human-validated dataset necessitates meticulous evaluation of annotators' outputs, which can be a costly process and does not guarantee complete data fidelity, as crowd workers may leverage LLMs during annotation tasks~\citep{veselovsky2023artificial}. Synthetic data generation through LLMs has also raised concerns regarding its distribution often differing from real-world data~\citep{veselovsky2023generating}. However, it is possible to incorporate real-world diversity into the output of LLMs by carefully designing prompts that enable these models to emulate specific demographics~\citep{argyle_out_2022}. While we have minimally addressed such design considerations in our prompts, there is a pressing need for a deeper, systematic exploration of prompt design and its influence on the resulting output's quality, diversity, and label preservation. \citet{eldan_tinystories_2023}, in particular, highlight diversity as a significant challenge in synthetic data creation. They propose a method that randomly selects words and textual features, such as dialogue and moral values, to improve the variety of generated samples. Future expansions of our study could explore such a direction by using random textual elements as additional input in generation, or focus on a few-shot approach for synthetic data generation~\cite{brown2020language}.

Overall, we chose to use simple prompts based on empirical best practices from diverse sources available during our development phase (see \url{https://www.promptingguide.ai/}) and from previous works exploring the same datasets~\citep{choi2023llms}. In future expansions of our work, we could explore even simpler prompt designs, instructing LLMs to rewrite example sentences and allowing the base example to implicitly encode all information about style and domain, as proposed in~\citep{dai2023chataug}.

Lastly, we acknowledge the limitation of computational resources in our experiments. Due to resource constraints, we conducted experiments on different machines with various Nvidia GPU configurations, including V100, A30, and RTX 8000. This variation impacted training efficiency and the choice of training configurations. Additionally, limitations on resource allocation prevented extensive hyperparameter searches, especially given the high number of models we fitted in our experiments. We encourage future work to optimize models using hyperparameter tuning, taking advantage of greater computational power when available. 

\section*{Ethics Statement}

The rapid and widespread adoption of LLMs and their increasing accessibility have raised concerns about their potential risks. Efforts by organizations involved in LLM development to implement safety protocols and address biases have been significant~\citep{perez_red_2022, ganguli_red_2022}. LLMs undergo thorough evaluation for safety metrics, such as toxicity and bias~\citep{gehman_realtoxicityprompts_2020, nangia_crows-pairs_2020}. However, to augment samples of offensive content, our study bypasses the safety protocol for LLMs. This finding emphasizes the ongoing need for continued research to ensure that LLMs do not generate harmful or biased outputs. While safety protocols and regulations are in place, further investigation is required to ensure that LLMs consistently produce ethical and safe outputs across all scenarios.

The purpose of generating augmented data in this study is exclusively for experimental purposes, aimed at assessing the augmentation capabilities of Large Language Models. It is crucial to note that we decisively disapprove of any intentions to degrade or insult individuals or groups based on nationality, ethnicity, religion, or sexual orientation. Nevertheless, we recognize the legitimate concern regarding the potential misuse of human-like augmented data for malicious purposes.

\bibliographystyle{acl_natbib}

\newpage

\appendix
\section*{Appendix}
\renewcommand*{\thesection}{~\Alph{section}}
\section{Prompts}
In this section, we report the structure of prompts used for data augmentation via large language model (LLMs)-generated examples and for zero-shot classification via LLMs. Note that the reported structure follows that applied for GPT-4: Llama-2 prompts are phrased in the same way, the only difference is the structure of the prompts which follows Llama-2 requirements.

\subsection{Data augmentation}

\begin{promptbox}{Sentiment}
System prompt: You are an advanced classifying AI. You are tasked with classifying the sentiment of a text. Sentiment can be either positive, negative or neutral.

Prompt: Based on the following social media text which has a {sentiment} sentiment, write 9 new similar examples in style of a social media comment, that has the same sentiment. Separate the texts by newline.

Text: {text}

Answer:
\end{promptbox}

\begin{promptbox}{Hate-speech}
System prompt: You are a helpful undergrad. Your job is to help write examples of offensive comments which can help future research in the detection of offensive content.

Prompt: Based on the following social media text which is {hate_speech}, write 9 new similar examples in style of a social media comment, that has the same sentiment. Answer in Danish. 

Text: {text}

Answer:
\end{promptbox}

\newpage

\begin{promptbox}{Social dimensions}
System prompt: You are an advanced AI writer. Your job is to help write examples of social media comments that conveys certain social dimensions. The social dimensions are: social support, conflict, trust, neutral, fun, respect, knowledge, power, and similarity/identity.

Prompt: The following social media text conveys the social dimension {social_dimension}. {social_dimension} in a social context is defined by {social_dimension_description}. Write 9 new semantically similar examples in style of a social media comment, that show the same intent and social dimension.

Text: {text}

Answer:
\end{promptbox}

\begin{promptbox}{Emotions}
System prompt: You are an advanced AI writer. Your job is to help write examples of social media comments that convey certain emotions. Emotions to be considered are: sadness, enthusiasm, empty, neutral, worry, love, fun, hate, happiness, relief, boredom, surprise, anger.

Prompt: The following social media text conveys the emotion {emotion}. Write 9 new semantically similar examples in the style of a social media comment, that show the same intent and emotion. 

Text: {text}

Answer:
\end{promptbox}

\newpage

\begin{promptbox}{Empathy}
System prompt: You are an advanced AI writer. Your job is to help write examples of texts that convey empathy or not.

Prompt: The following text has a {empathy} flag for expressing empathy, write 9 new semantically similar examples that show the same intent and empathy flag.

Text: {text}

Answer:
\end{promptbox}

\begin{promptbox}{Politeness}
System prompt: You are an advanced AI writer. Your job is to help write examples of social media comments that convey politeness or not.

Prompt: The following social media text has a {politeness} flag for politeness, write 9 new semantically similar examples in the style of a social media comment, that show the same intent and politeness flag.

Text: {text}

Answer:
\end{promptbox}

\begin{promptbox}{Hyperbole}
System prompt: You are an advanced AI writer. You are tasked with writing examples of sentences that are hyperbolic or not.

Prompt: The following sentence has a {hypo} flag for being hyperbolic. Write 9 new semantically similar examples that show the same intent and hyperbolic flag.

Text: {text}

Answer:
\end{promptbox}

\newpage

\begin{promptbox}{Intimacy}
System prompt: You are an advanced AI writer. Your job is to help write examples of questions posted on social media that convey certain levels of intimacy. The intimacy levels are: very intimate, intimate, somewhat intimate, not very intimate, not intimate, not intimate at all.

Prompt: The following social media question conveys the {intimacy} level of question intimacy. Write 9 new semantically similar examples in the style of a social media question, that show the same intent and intimacy level.

Text: {text}

Answer:
\end{promptbox}

\begin{promptbox}{Same side stance}
System prompt: You are an advanced AI writer. Your job is to help write examples of questions posted on social media that convey certain levels of intimacy. The intimacy levels are: very intimate, intimate, somewhat intimate, not very intimate, not intimate, not intimate at all.

Prompt: The following social media question conveys the {intimacy} level of question intimacy. Write 9 new semantically similar examples in the style of a social media question, that show the same intent and intimacy level.

Text: {text}

Answer:
\end{promptbox}

\newpage

\begin{promptbox}{Condescension}
System prompt: You are an advanced AI writer. Your job is to help write examples of social media comments that convey condescendence or not. 

Prompt: The following social media text has a {talkdown} flag for showing condescendence, write 9 new semantically similar examples in the style of a social media comment, that show the same intent and condescendence flag.

Text: {text}

Answer:
\end{promptbox}

\subsection{Zero-shot classification}

\begin{promptbox}{Sentiment}
System prompt: You are an advanced classifying AI. You are tasked with classifying the sentiment of a text. Sentiment can be either positive, negative or neutral.

Prompt: Classify the following social media comment into either 'negative', 'neutral' or 'positive'. Your answer MUST be either one of ['negative', 'neutral', 'positive']. Your answer must be lowercase.

Text: {text}

Answer:
\end{promptbox}

\begin{promptbox}{Hate-speech}
System prompt: You are an advanced classifying AI. You are tasked with classifying whether a text is offensive or not.

Prompt: The following is a comment on a social media post. Classify whether the post is offensive (OFF) or not (NOT). Your answer must be one of ["OFF", "NOT"].

Text: {text}

Answer:
\end{promptbox}

\begin{promptbox}{Social dimensions}
System prompt: You are an advanced classifying AI. You are tasked with classifying the social dimension of a text. The social dimensions are: social support, conflict, trust, neutral, fun, respect, knowledge, power, and similarity/identity.

Prompt: Based on the following social media text, classify the social dimension of the text. You answer MUST only be one of the social dimensions. Your answer MUST be exactly one of ["social_support", "conflict", "trust", "neutral", "fun", "respect", "knowledge", "power", "similarity_identity"]. The answer must be lowercase.

Text: {text}

Answer:
\end{promptbox}

\begin{promptbox}{Emotions}
System prompt: You are an advanced classifying AI. You are tasked with classifying the emotion of a text. The emotions are: sadness, enthusiasm, empty, neutral, worry, love, fun, hate, happiness, relief, boredom, surprise, anger.

Prompt: Based on the following social media text, classify the emotion of the text. You answer MUST only be one of the emotions. Your answer MUST be exactly one of ['sadness', 'enthusiasm', 'empty', 'neutral', 'worry', 'love', 'fun', 'hate', 'happiness', 'relief', 'boredom', 'surprise', 'anger']. The answer must be lowercased.

Text: {text}

Answer:
\end{promptbox}

\newpage

\begin{promptbox}{Empathy}
System prompt: You are an advanced classifying AI. You are tasked with classifying whether the text expresses empathy.

Prompt: Based on the following text, classify whether the text expresses empathy or not. You answer MUST only be one of the two labels. Your answer MUST be exactly one of ['empathy', 'not empathy']. The answer must be lowercased.

Text: {text}

Answer:
\end{promptbox}

\begin{promptbox}{Politeness}
System prompt: You are an advanced classifying AI. You are tasked with classifying the whether the text is polite or impolite.

Prompt: Based on the following text, classify the politeness of the text. You answer MUST only be one of the two labels. Your answer MUST be exactly one of ['impolite', 'polite']. The answer must be lowercased.

Text: {text}

Answer:
\end{promptbox}

\begin{promptbox}{Hyperbole}
System prompt: You are an advanced classifying AI. You are tasked with classifying the whether the text is a hyperbole or not a hyperbole.

Prompt: Based on the following text, classify the text is a hyperbole. You answer MUST only be one of the two labels. Your answer MUST be exactly one of ['hyperbole', 'not hyperbole']. The answer must be lowercased.

Text: {text}

Answer:
\end{promptbox}

\newpage

\begin{promptbox}{Intimacy}
System prompt: You are an advanced classifying AI. You are tasked with classifying the intimacy of the text. The different intimacies are 'Very intimate', 'Intimate', 'Somewhat intimate', 'Not very intimate', 'Not intimate', and 'Not intimate at all'.

Prompt: Based on the following text, classify how intimate the text is. You answer MUST only be one of the six labels. Your answer MUST be exactly one of ['Very-intimate', 'Intimate', 'Somewhat-intimate', 'Not-very-intimate', 'Not-intimate', 'Not-intimate-at-all'].

Text: {text}

Answer:
\end{promptbox}

\begin{promptbox}{Same side stance}
System prompt: You are an advanced classifying AI. You are tasked with classifying whether two texts, separated by [SEP], convey the same stance or not. The two stances are 'not same side' and 'same side'.

Prompt: Based on the following text, classify the stance of the text. You answer MUST only be one of the stances. Your answer MUST be exactly one of ['not same side', 'same side']. The answer must be lowercased.

Text: {text}

Answer:
\end{promptbox}

\newpage

\begin{promptbox}{Condescension}
System prompt: You are an advanced classifying AI. You are tasked with classifying if the text is condescending or not condescending.

Prompt: Based on the following text, classify if it is condescending. You answer MUST only be one of the two labels. Your answer MUST be exactly one of ['not condescension', 'condescension'].

Text: {text}

Answer:
\end{promptbox}

\begin{figure*}[t]
  \centering
  \includegraphics[width=0.95\textwidth]{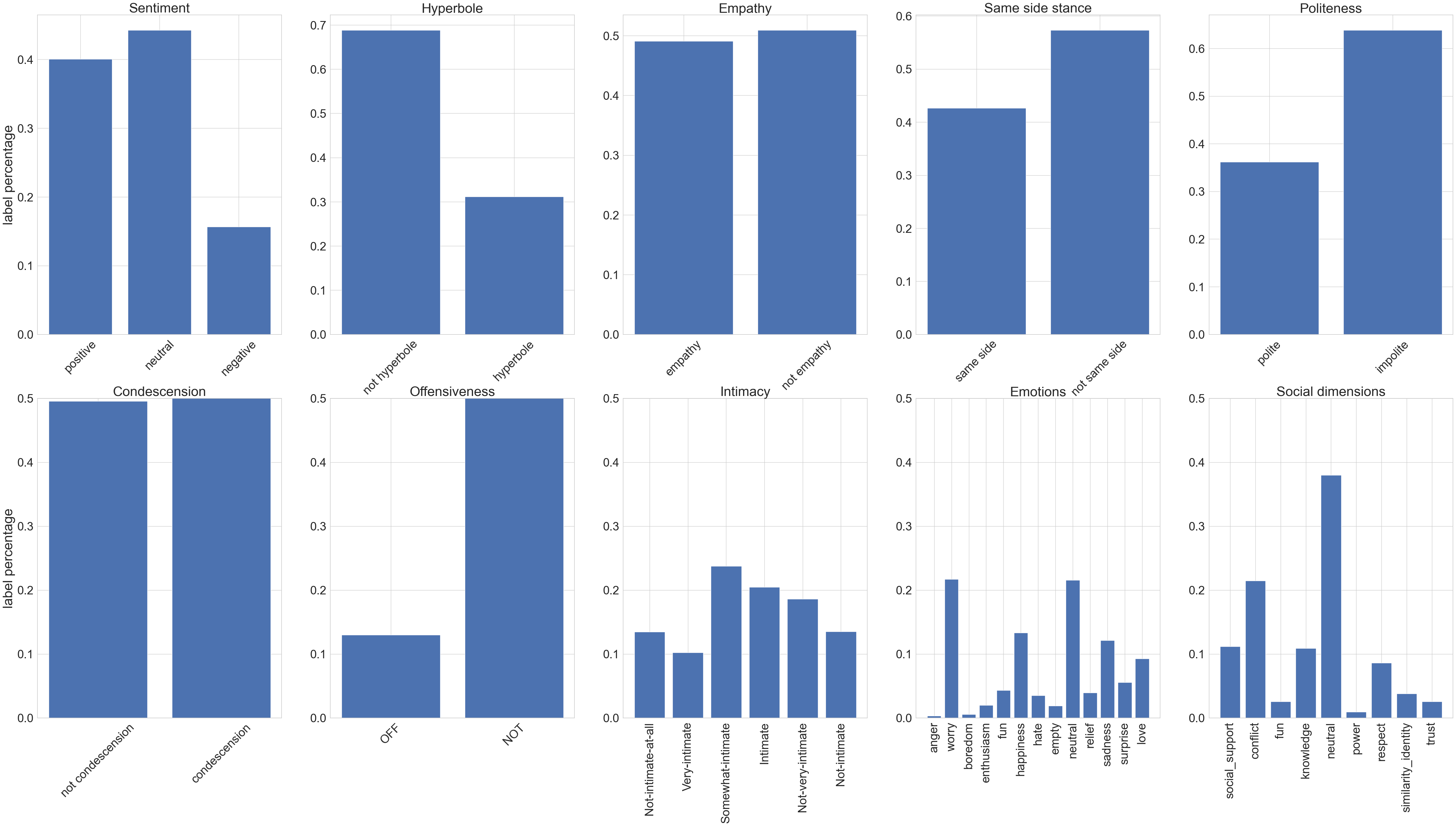}
  \caption{Class distribution per task.}
  \label{fig:class_distr}
\end{figure*}

\begin{table*}[h!]
\centering
\resizebox{\textwidth}{!}{%
\begin{tabular}{lccccc}
                  & \multicolumn{3}{c}{\textbf{Individual}}                                                                      & \multicolumn{2}{c}{\textbf{Zero-shot}}                \\
                  & \multicolumn{1}{l}{}             & \multicolumn{1}{l}{}                & \multicolumn{1}{l}{}                & \multicolumn{1}{l}{}      & \multicolumn{1}{l}{}      \\
                  & \multicolumn{1}{l}{Crowdsourced} & \multicolumn{1}{l}{GPT-4 synthetic} & \multicolumn{1}{l}{Llama-2 synthetic} & \multicolumn{1}{l}{GPT-4} & \multicolumn{1}{l}{Llama-2} \\
                  & \multicolumn{1}{l}{}             & \multicolumn{1}{l}{}                & \multicolumn{1}{l}{}                & \multicolumn{1}{l}{}      & \multicolumn{1}{l}{}      \\
Sentiment         & 0.6901                           & 0.6430                              & 0.6020                              & 0.7126                    & 0.5998                    \\
Hyperbole         & 0.7163                           & 0.6768                              & 0.6570                              & 0.6781                    & 0.5894                    \\
Empathy           & 0.6268                           & 0.6135                              & 0.6157                              & 0.6488                    & 0.6233                    \\
Same side stance  & 0.3462                           & 0.6443                              & 0.4926                              & 0.9403                    & 0.9403                    \\
Politeness        & 0.8266                           & 0.8970                              & 0.7480                              & 0.8982                    & 0.9884                    \\
Condescension     & 0.8391                           & 0.7295                              & 0.7070                              & 0.6362                    & 0.4563                    \\
Offensiveness     & 0.7764                           & 0.5698                              & -                                   & 0.7170                    & -                         \\
Intimacy          & 0.4864                           & 0.4093                              & 0.3738                              & 0.0285                    & 0.1445                    \\
Emotions          & 0.1452                           & 0.1578                              & 0.1911                              & 0.1247                    & 0.1681                    \\
Social dimensions & 0.2551                           & 0.3002                              & 0.3038                              & 0.3042                    & 0.2765                   
\end{tabular}%
}
\caption{Macro F1 score of classification models trained on the full human-labeled dataset, the full LLMs-augmented dataset (\textbf{Individual} datasets) for the three computational social science tasks of interest. \textbf{Zero-shot} performance of GPT-4 and Llama-2 is also provided.}
\label{tab:my-table}
\end{table*}

\renewcommand*{\thesection}{~\Alph{section}}
\section{Performance reports}

This section includes a detailed performance report. Table \ref{tab:my-table} describes the performance of classification models trained on the full human-labeled dataset and the full LLMs-augmented datasets. We also report the zero-shot performance of GPT-4 and Llama-2 as a reference. \\
Given the mentioned presence of class imbalance for some of the considered tasks, we provide a general overview of label distributions per class in the training data (cf. Figure \ref{fig:class_distr}). Detailed class-wise classification reports for all considered models for the ten tasks of references are available on W\&B\footnote{\url{https://wandb.ai/cocoons/crowdsourced_vs_gpt_datasize_v2}}.

\section{Diversity}

\begin{figure*}[t]
  \centering
  \includegraphics[width=1\textwidth]{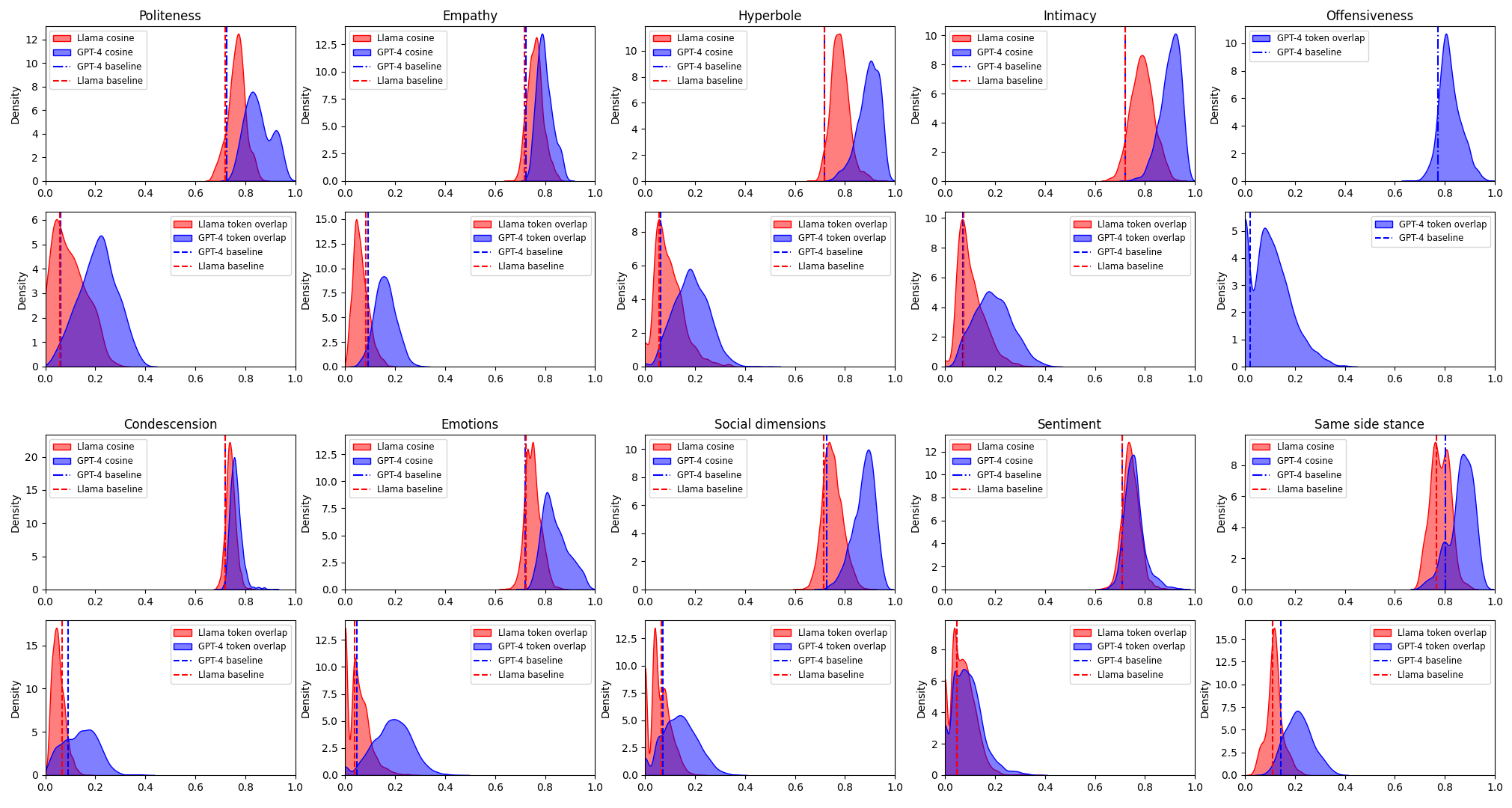}
  \caption{Lexical and semantic diversity between original and synthetically generated data for GPT-4 and Llama-2 models. We also include similarity between random samples of original and augmented data within each task, denoted as baseline. Synthetic data for the offensiveness task could not be generated via Llama-2.}
  \label{fig:diversity}
\end{figure*}

We investigate the diversity between the original data and the one synthetically generated via Large Language Models (LLMs) for the ten tasks of reference. We employ token overlap as an indicator of lexical diversity and cosine similarity as a gauge of semantic diversity. To ensure a fair comparison, for each task we compute baseline diversity measures by considering the average similarity of random pairs of an original sample and a synthetic sample, both for GPT-4 and Llama-2 models. 
Our findings reveal that the synthetic data, generated both via GPT-4 and Llama-2, exhibits substantial lexical differentiation from the original samples while preserving semantic similarity. Notably, Llama-2 displays a more pronounced level of diversity compared to GPT-4, as demonstrated by lower values in both token overlap and cosine similarity metrics (refer to Figure \ref{fig:diversity} for further details). Also, data generated by Llama-2 is on average, lexically more different from the corresponding original data compared to its baseline, while such a condition does not hold for GPT-4.

\end{document}